\documentclass[a4paper,fleqn]{cas-dc}

\makeatletter
\renewcommand{\fnum@figure}{Fig. \thefigure.\@gobble}
\makeatother

\usepackage[justification=centering]{caption} 
\usepackage[numbers]{natbib}
\usepackage{amsmath}
\usepackage{amsmath, bm}
\usepackage{bbding}
\usepackage{pifont}
\usepackage{amsfonts}
\usepackage{dsfont}
\usepackage{color}
\usepackage{bbm}
\usepackage{textcomp}

\begin{document}
\begin{sloppypar}
\let\WriteBookmarks\relax
\def\floatpagepagefraction{1}
\def\textpagefraction{.001}

\title[mode = title]{Twin Deformable Point Convolutions for Airborne Laser Scanning Point Cloud Classification}


\author[1,2,3,4]{Yong-Qiang Mao} \ead{maoyongqiang19@mails.ucas.ac.cn}
\author[1,2,3,4]{Hanbo Bi}\ead{bihanbo21@mails.ucas.ac.cn}
\author[1,2,3,4]{Xuexue Li} \ead{lixuexue20@mails.ucas.ac.cn}
\author[1,2]{Kaiqiang Chen}\fnmark[*] \ead{chenkaiqiang14@mails.ucas.ac.cn}
\author[1,2]{Zhirui Wang}\ead{zhirui1990@126.com}
\author[1,2,3,4]{Xian Sun} \ead{sunxian@aircas.ac.cn}
\author[1,2,3,4]{Kun Fu} \ead{kunfuiecas@gmail.com}

\address[1]{Aerospace Information Research Institute, Chinese Academy of Sciences, Beijing 100190, China}
\address[2]{Key Laboratory of Network Information System Technology (NIST), Aerospace Information Research Institute, Chinese Academy of Sciences, Beijing 100190, China}
\address[3]{University of Chinese Academy of Sciences, Beijing 100190, China}
\address[4]{School of Electronic, Electrical and Communication Engineering, University of Chinese Academy of Sciences, Beijing 100190, China}

\tnotetext[1]{Corresponding author. \\
This work was supported by the National Natural Science Foundation
of China under Grant 62331027 and Grant 62076241, and supported by the Strategic Priority Research Program of the Chinese Academy of Sciences, Grant No. XDA0360300. }

\begin{abstract}[abstract]
Thanks to the application of deep learning technology in point cloud processing of the remote sensing field, point cloud segmentation has become a research hotspot in recent years, which can be applied to real-world 3D, smart cities, and other fields. Although existing solutions have made unprecedented progress, they ignore the inherent characteristics of point clouds in remote sensing fields that are strictly arranged according to latitude, longitude, and altitude, which brings great convenience to the segmentation of point clouds in remote sensing fields. To consider this property cleverly, we propose novel convolution operators, termed Twin Deformable point Convolutions (TDConvs), which aim to achieve adaptive feature learning by learning deformable sampling points in the latitude-longitude plane and altitude direction, respectively. First, to model the characteristics of the latitude-longitude plane, we propose a Cylinder-wise Deformable point Convolution (CyDConv) operator, which generates a two-dimensional cylinder map by constructing a cylinder-like grid in the latitude-longitude direction, and then performs adaptive feature sampling on the cylinder map by deformable offset learning. Furthermore, to better integrate the features of the latitude-longitude plane and the spatial geometric features, we perform a multi-scale fusion of the extracted latitude-longitude features and spatial geometric features, and realize it through the aggregation of adjacent point features of different scales. In addition, a Sphere-wise Deformable point Convolution (SpDConv) operator is introduced to adaptively offset the sampling points in three-dimensional space by constructing a sphere grid structure, aiming at modeling the characteristics in the altitude direction. Experiments on existing popular benchmarks conclude that our TDConvs achieve the best segmentation performance, surpassing the existing state-of-the-art methods. The code is available on \url{https://github.com/WingkeungM/TDConvs}.
\end{abstract}

\maketitle

\section{Introduction}\label{Introduction}
In recent years, with the rapid advancement of deep learning technology and the explosive growth of remote sensing satellite data, the interpretation methods of remote sensing data have achieved unprecedented development.
Especially the research on 3D remote sensing data has become increasingly popular. As the main sensor for collecting point cloud data in the field of remote sensing, airborne laser scanning (ALS) technology,  has gradually entered the field of vision of researchers. Among them, ALS point cloud segmentation has received a lot of attention.

\begin{figure*}[htb]
    \setlength{\abovecaptionskip}{1pt}
    \centering
    \includegraphics[width=1.0\linewidth]{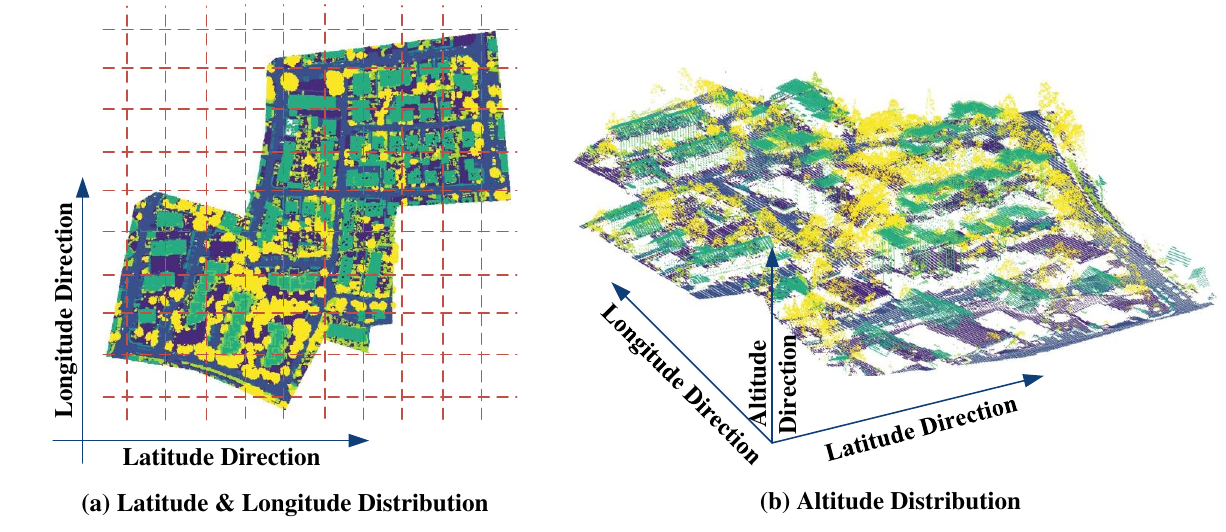}
    \caption{ALS point clouds are acquired by the aircraft, therefore, these point clouds are strictly arranged according to the latitude and longitude direction and the altitude direction. \textbf{(a) Latitude and Longitude Distribution.} The instance objects in the large-scale point cloud are strictly arranged in fixed latitude and longitude positions. \textbf{(b) Altitude Distribution.} Different instance objects have different altitudes, which is one of the inherent characteristics of large-scale city-level point clouds.}
    \label{Fig1-motivation}
\end{figure*}

Thanks to the development of various deep learning technologies, ALS point cloud segmentation technology has evolved from traditional support vector machines\cite{suykens1999least}, random forests\cite{svetnik2003random}, and other methods\cite{hastie2009multi} to the current MLP-based methods\cite{qi2017pointnet,qi2017pointnet++}, graph-based methods\cite{wang2019graph,mao2022beyond}, and point convolution-based methods\cite{li2018pointcnn,wu2019pointconv,thomas2019kpconv}. 
MLP-based methods\cite{qi2017pointnet,qi2017pointnet++} utilize MLP operations for point-by-point feature learning in point sets, such as PointNet\cite{qi2017pointnet} and PointNet++\cite{qi2017pointnet++}. However, it has certain limitations in learning the internal features of point sets. Researchers have proposed graph-based methods\cite{wang2019graph,mao2022beyond} and point convolution-based methods\cite{li2018pointcnn,jiang2018pointsift,wu2019pointconv,thomas2019kpconv}. Graph-based methods aim to perform deep feature extraction of point sets through graph feature learning by modeling point sets as graph structures. The method based on point convolution is inspired by the success of convolution in images and proposes a variety of convolution methods for point sets to perform precise feature extraction.
In ALS point cloud processing, Wen et al.\cite{wen2020directionally} propose a direction-constrained fully convolutional neural network D-FCN, which extracts local representative features of 3D point sets from the projected 2D receptive field through direction-constrained point convolution. Li et al.\cite{li2020dance} introduce a density-aware convolution module that uses point-wise density to reweight the learnable weights of the convolution kernel. Li et al.\cite{li2020geometry1} propose a geometric attention network composed of geometry-aware convolutions, dense hierarchies, and elevation attention modules to effectively embed features.

However, these methods ignore a problem, that is, although the ALS point cloud is disordered in data form, its spatial distribution is strictly arranged according to latitude (x coordinate), longitude (y coordinate), and altitude (z coordinate), which brings great convenience to the point cloud segmentation of large scenes. 
Existing methods often focus on the extraction of representative features of point clouds, ignoring the 
specific geographic information arrangement characteristics of ALS point clouds according to latitude, longitude, and altitude. 

On the one hand, as shown in Fig.~\ref{Fig1-motivation}(a), unlike the point cloud randomly arranged in 3D space in the visual scene, the ALS point cloud is strictly arranged according to latitude and longitude. 
If the point cloud of the visual scene is projected to any two-dimensional plane in the three-dimensional space, we cannot get a reasonable plane to better characterize the characteristics of the point cloud scene.
However, from an aerial perspective, latitude and longitude are important physical properties of the acquired lidar point cloud. Projecting the ALS point cloud to the latitude-longitude plane can obtain the lidar bird's-eye-view, and the bird's-eye-view plane is the most intuitive plane to represent the remote sensing scene. This is also a unique feature of aerial lidar point clouds compared to visual scenes. Therefore, if it can understand the positional relationship of instances on the latitude-longitude plane, that is, the bird's-eye-view plane, it will help the network to better distinguish densely packed and adjacent objects, to perform better segmentation. 
To better learn this characteristic, we propose a novel convolution operator called Cylinder-wise Deformable point Convolution (CyDConv), which adaptively samples the features in the latitude-longitude plane by constructing deformable reference points to achieve feature fusion.

On the other hand, in the altitude direction, the objects are also arranged according to their actual spatial altitude positions, as shown in Fig.~\ref{Fig1-motivation}(b). For example, cars are typically located in lower-altitude distribution areas, while powerlines are located in higher-altitude areas. Making full use of altitude information for learning can clearly distinguish such objects with typical altitude differences.
To solve this problem, we propose a novel Sphere-wise Deformable point Convolution (SpDConv) operator, which aggregates point clouds into regions in 3D space by constructing sphere structures and constructs 3D reference points between spheres to sample sphere features for the perception of altitude features.

To the end, we propose a Twin Deformable point Convolutions (TDConvs) method to enable the modeling of geographic information in ALS point cloud segmentation.
First, our method follows the classic encoder-decoder structures for high-level feature extraction and resolution recovery, respectively. Second, in the encoder stage, each layer of our encoder is rooted in our proposed Cylinder-wise Deformable point Convolution (CyDConv) to realize the modeling and learning of latitude and longitude features in the geographic information characteristics of ALS point clouds. Furthermore, to better model the altitude features in geographic information characteristics, we introduce our Sphere-wise Deformable point Convolution (SpDConv) in the skip-connection process of the encoder and decoder. With both CyDConv and SpDConv embedded in the network, our TDConvs method enables the modeling and efficient learning of geographic information features of ALS point clouds in remote sensing scenes.

The main contributions of our work can be summarized as follows:
\begin{itemize}
    \item We proposed TDConvs, which explicitly models the geographic information of the ALS point cloud to boost the performance of point cloud segmentation.
    \item We built a novel Cylinder-wise Deformable point Convolution (CyDConv) operator, to model the latitude and longitude characteristics through an adaptive feature sampling manner according to the two-dimensional reference points of the latitude-longitude plane.
    \item We formulate a novel Sphere-wise Deformable point Convolution (SpDConv) operator, to explicitly learn the altitude features in the skip connection structure between encoder and decoder, which is complemented by an adaptive spatial sampling method inside a sphere structure.
    \item Extensive experiments on popular benchmarks show that our proposed TDConvs achieve new state-of-the-art performance, significantly outperforming existing solutions.
\end{itemize}

\section{Related Work}\label{sec:Related Work}
\subsection{Classical Machine Learning-Based Methods}
The specific process of the point cloud semantic segmentation method based on artificial feature learning is as follows: first, the artificially designed features of the point cloud data are extracted, and then the extracted semantic information is semantically classified using commonly used classifiers.
In recent years, the artificially designed features of point cloud data have mainly focused on the structural geometric features inside the point cloud\cite{lalonde2006natural,munoz2009onboard,belongie2002shape}, 3D descriptor of the 3D shape design of point cloud\cite{frome2004recognizing}, histogram of point cloud\cite{rusu2010fast,tombari2010unique}, etc.
The structural geometric characteristics of point clouds focus on modeling the internal characteristics of point sets, including the structure of point clouds, normal vectors, and structural characteristics of points, lines, and surfaces.
Specifically, by utilizing artificially designed local structural geometric features, Lalonde et al.\cite{lalonde2006natural} focused on three types of objects: ``dispersed" such as grass and tree crowns, ``linear" such as wires and branches, and ``surface" such as ground and rocks. Shape descriptors are also an effective modeling method for point cloud features. To detect point clouds of tree categories in urban areas, Monnier et al.\cite{monnier2012trees} used descriptors of local structural features for point cloud collections based on fixed neighborhoods. 
The histogram of point cloud can be used to describe point cloud features through computational methods.
Ruse et al.\cite{rusu2010fast} designed feature descriptions by encoding geometric and viewpoint features of point cloud collections, and proposed a histogram modeling method of viewpoint features, which was extremely robust and promotes practical application.
After the point cloud manual feature extraction, the features are input into the mainstream classifier based on machine learning methods. Typical examples include Gaussian mixture model\cite{bishop2006pattern}, support vector machine\cite{colgan2012mapping,garcia2015evolutionary}, Adaboost\cite{lodha2007aerial}, random forest \cite{Blomley_and_Weinmann_2017,niemeyer2012conditional}, etc.

However, the classical manual extraction methods are inefficient and cannot be applied to various point cloud scenarios.

\subsection{Point Cloud Segmentation Based on Deep Learning}
The rapid development of deep learning has promoted the replacement of manual feature design methods by automatic point cloud feature extraction\cite{wang2019dynamic,thomas2019kpconv, qi2017pointnet}. Among them, point-based methods are the most favored by researchers among many methods. 
The point-based methods\cite{qi2017pointnet,qi2017pointnet++,kolodiazhnyi2024oneformer3d,zhan2023fa,geng20233dgraphseg,yin2023dcnet} refer to directly processing the point cloud without any data preprocessing operation, and then achieving the classification and segmentation tasks of the input point set by learning the deep semantic features of the point set. It mainly includes MLP-based methods\cite{qi2017pointnet,qi2017pointnet++,liu2019relation,wen2020directionally}, graph-based methods\cite{wang2019graph,shi2020point,te2018rgcnn}, and point convolution-based methods\cite{li2018pointcnn,li2020geometry1,wu2019pointconv,thomas2019kpconv,hu2020randla}. 
MLP-based methods\cite{qi2017pointnet,qi2017pointnet++,liu2019relation,wen2020directionally} are the earlier deep convolutional neural network methods for processing point cloud data. 
PointNet\cite{qi2017pointnet} is the pioneering work of this type of method. It first uses multiple multi-layer perceptrons with shared weights to learn point-by-point high-dimensional features, and then designs a symmetric function based on the maximum pooling function to solve the disorder problem of point cloud. 
PointNet++\cite{qi2017pointnet++} models local relationships through the three processes of sampling, grouping, and feature learning inside the point cloud. 
Point sets are the discrete structure, and there is a certain connection relationship between the points. This is a natural graph structure data form. Therefore, some researchers have introduced graph structure modeling into point cloud learning and achieved excellent results.
Landrieu et al.\cite{landrieu2018large} earlier proposed a point cloud convolutional neural network SPGraph based on graph convolution, which used graph convolution to learn high-level semantic features of the graph structure, and ultimately achieved point-by-point semantic segmentation. Although traditional convolution has been successful in feature extraction of 2D raster image data, it cannot be directly transferred to point cloud processing problems due to the unique properties of point cloud data.
Therefore, Li et al.\cite{li2018pointcnn} tried to weight the features of the input point set and use a multi-layer perceptron to learn the X transformation to achieve point-by-point weighted summation in the neighborhood.
In addition, Li et al.\cite{li2020geometry1} proposed a geometric attention network composed of geometry-aware convolution, dense hierarchical architecture, and elevation attention module, and trained in an end-to-end manner to solve the three problems (geometric instances, extreme scale changes, and large differences in elevation) of remote sensing point clouds.

Point-based methods have brought rapid progress to point cloud processing, which has promoted the application of point cloud learning in various fields. However, in remote sensing scenarios, current methods ignore that point cloud data are strictly arranged according to longitude, latitude, and altitude, which is different from the disorder of the data form. Therefore, this paper focuses on the explicit modeling of the longitude, latitude, and altitude of point sets.

\begin{figure*}[ht]
    \setlength{\abovecaptionskip}{1pt}
    \centering
    \includegraphics[width=0.8\linewidth]{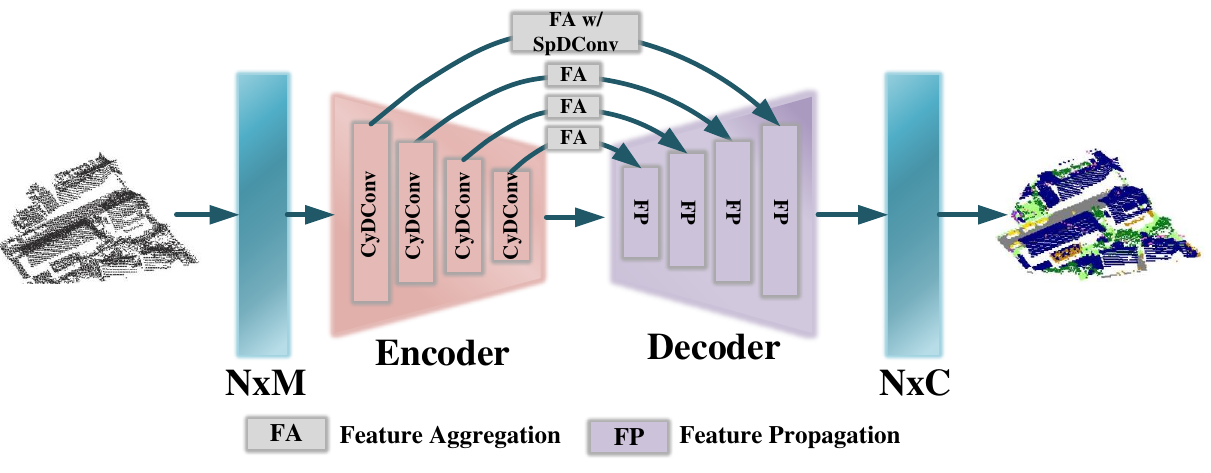}
    \caption{\textbf{Overall Architecture.} Our TDConvs adopt the popular Encoder-Decoder structure as the basic component. Each feature extraction layer in the encoder part is the proposed Cylinder-wise Deformable point Convolution (CyDConv), and each layer in the decoder part is a feature forward propagation layer. The encoder and decoder are connected as feature aggregation layers with our proposed Sphere-wise Deformable point Convolution (SpDConv) at the highest resolution layer. The dimension of input point sets is $\rm N\times M$, where $\rm M$ is the dimension of point features. The dimension of output point sets is $\rm N\times C$, where $\rm C$ is the number of output categories.}
    \label{Fig2-framework}
\end{figure*}

\subsection{Geospatial Modeling in Point Cloud Segmentation}
In ALS point cloud learning, fine modeling of geographic geometric features is one of the effective means of large scene point cloud processing. Among them, GADH-Net\cite{li2020geometry1} cleverly uses the features of the ALS point cloud that are different from the point cloud in visual scenes, that is, clear geographic altitude information, and builds an elevation attention module to learn height information from the lowest-level features and then embeds it into the final decoded features which are used for point cloud segmentation. 
Chen et al.\cite{chen2021pointnet++} propose an elevation and distance-based interpolation method to perform interpolation of point cloud features between different levels to achieve geographic altitude modeling. 
Ni et al.\cite{ni2017classification} define four elevation-based features which are relative elevation, elevation variance, elevation difference, and normalized elevation, to model geographic altitude features.

However, these methods only consider the altitude characteristics of ALS point clouds, ignoring that these point clouds are arranged strictly according to the characteristics of longitude and latitude. To this end, we propose novel convolution operators based on longitude-latitude modeling and altitude modeling to realize the geographic geometric feature learning of ALS point cloud.

\section{Proposed Method}\label{sec:Proposed Method}
\subsection{Method Overview}\label{sec:Method Overview}
To better learn the geographical distribution characteristics of the ALS point cloud that is strictly arranged according to latitude, longitude, and altitude, we propose Twin Deformable point Convolutions (TDConvs) to realize the feature offset learning of point cloud in latitude, longitude, and altitude, to fully embed these geospatial information into the features of point sets.

As shown in Fig.~\ref{Fig2-framework}, the flowchart of our entire framework is given. 
Given the sampled point set, the encoder and decoder are executed sequentially. In addition, there is a skip-connection layer between each layer of the encoder and decoder for feature fusion. We embed the learning of longitude and latitude information into Cylinder-wise Deformable point Convolution (CyDConv), and the learning of altitude information into Sphere-wise Deformable point Convolution (SpDConv). CyDConv is performed in each feature extraction layer of the encoder, and SpDConv is performed in the highest resolution feature aggregation layer. The decoder is rooted in the feature propagation, which is proposed in PointNet++\cite{qi2017pointnet++}.

\subsection{Cylinder-wise Deformable Point Convolution}
Since the ALS point cloud is strictly arranged according to latitude and longitude, it is necessary to make better use of this distribution. Existing methods focus on the extraction of fine point cloud geometric features, ignoring the signification of geospatial space arrangement, for which we propose a novel Cylinder-wise Deformable point Convolution (CyDConv) to learn the difference of characteristics of point clouds in the latitude and longitude direction. Below we will introduce the implementation process of our CyDConv.

\begin{figure*}[htb]
	\setlength{\abovecaptionskip}{1pt}
	\centering
	\includegraphics[width=1.0\linewidth]{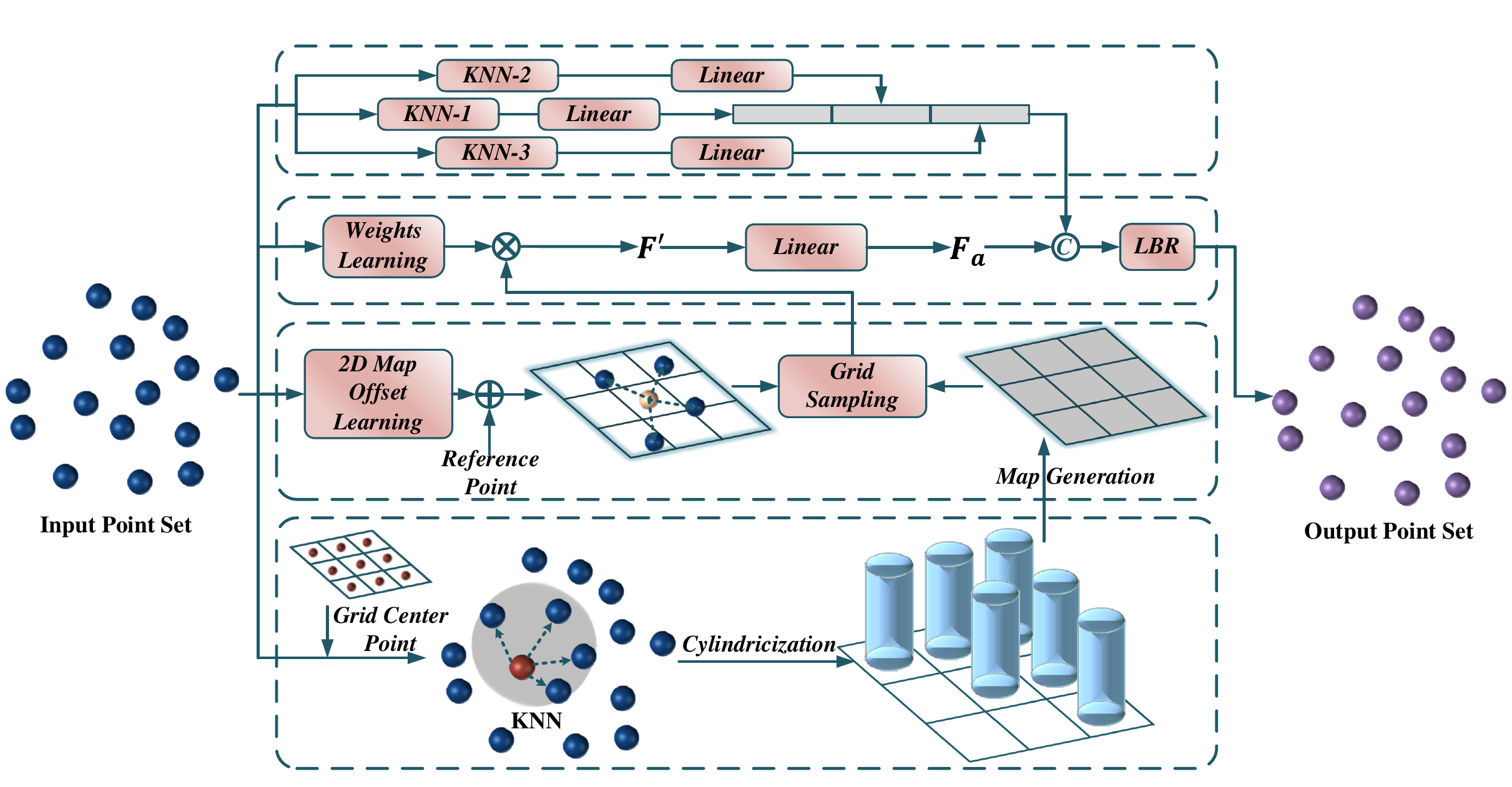}
	\caption{The architecture diagram of our proposed Cylinder-wise Deformable point Convolution (CyDConv). `LBR' means the Linear-BN-ReLU operation. The parts from top to bottom in the figure represent Multi-Nearest Neighbor Feature Learning (MFL), Map Feature Sampling and Aggregation, Map Offset Learning, and Cylindricization and Map Generation processes, respectively.}
	\label{Fig3-CyDConv}
\end{figure*}

\textbf{Cylindricization and Map Generation.}
Specifically, given the input point set $\mathbf{P}\in \mathbb{R}^{\rm N\times 3}$ and its feature $\mathbf{F}\in \mathbb{R}^{\rm N\times M}$, we first normalize the coordinates of the input point set $\mathbf{P}$, and then convert the point cloud into a cylinder structure at the top view, called cylindricization. Mathematically, we first construct the center points through the grid arrangement as shown in the bottom of Fig.~\ref{Fig3-CyDConv}, which matches each cylinder of the top view, as 
\begin{equation} 
    \rm \mathbf{C}_{cet} = \rm GridArrange(H_c, W_c)
\end{equation}
where $\rm\mathbf{C}_{cet}\in \mathbb{R}^{\rm H_c\times W_c\times 2}$ is the grid center point corresponding to each cylinder. $\rm GridArrange$ is the Grid Arrangement operation. Specifically, the Grid Arrangement operation is to generate certain grids according to a certain step size. 
$\rm H_c$ and $\rm W_c$ are the height size and width size of the cylinder map. 
Based on the center points $\rm\mathbf{C}_{cet}$, the cylindricization process is implemented by the cylinder with the grid point as the center and $\rm R_c$ as the base radius, as
\begin{equation}
    \rm \mathbf{\Omega}_{j} = \rm \bigcup \{\mathbf{F}_i\},\ \ s.t.\ ||\mathbf{P}_{i;xy}-\mathbf{C}_{cet;j}|| < R_{c}
\end{equation}
where $\rm \mathbf{P}_{i; xy}$ is the projection of the point $\rm \mathbf{P}_i$ on the base plane where the center point $\rm \mathbf{C}_{cet;j}$ is located. $\rm \mathbf{C}_{cet;j}$ is the $\rm j$-th center point in the grids. 
The point feature corresponding to the point $\rm \mathbf{P}_i$ is $\rm \mathbf{F}_i$. 
According to the equation, we can get the index of the points where the projection of the base is located in the area centered on the grid point and radiused by $\rm R_c$. 
Through using the index, we can get the set $\rm \mathbf{\Omega}_{j}$ of features belonging to the points inside the cylinder.
Thus, the cylinder map $\rm \mathbf{M}$ can be generated by 
\begin{equation}
    \rm \mathbf{M}_{j} = mean(\mathbf{\Omega}_{j})
\end{equation}
where $\rm \mathbf{M}_{j}$ is the $\rm j$-th element of $\rm \mathbf{M}$. 

\textbf{Map Offset Learning.} As shown in Fig.~\ref{Fig3-CyDConv}, to model the relationship within point sets in the latitude and longitude directions, we learn the offset through the point coordinates and point features, as
\begin{equation}
    \rm \Delta^{c} = Linear(cat(\mathbf{P}, \mathbf{F}))
\end{equation}
where $\rm \Delta^{c}\in \mathbb{R}^{N_c\times K_c\times 2}$ is the offset in the latitude and longitude directions (the cylinder map). $\rm Linear$ and $\rm cat$ represent the linear layer and concatenate operation, respectively. Each point has corresponding $\rm K_c$ offset points in the cylinder map, and the coordinates of each offset point are $\rm x$ and $\rm y$ values.

\textbf{Map Feature Sampling and Aggregation.} After obtaining the offset points on the cylinder map of each point, feature sampling is performed through the grid sampling operation, which can be formulated as
\begin{equation}
    \rm \mathbf{F}^c_{o} = GridSampling(\mathbf{M}, \mathbf{P}^c_{ref}+\Delta^{c})
\end{equation}
where $\rm \mathbf{P}^c_{ref}$ is the initial reference points of the offset with random initialization. $\rm \mathbf{P}^c_{ref}+\Delta^{c}$ is the points after offsets. $\rm \mathbf{F}^c_{o}\in \mathbb{R}^{N_c\times K_c\times M}$ is the feature that is sampled by the points with offsets. 

In order to better aggregate the features $\rm \mathbf{F}^c_{o}$ that characterizes the latitude and longitude characteristics sampled from the cylinder map, we use the original points $\rm \mathbf{P}$ and point features $\rm \mathbf{F}$ for the learning of the weight matrix, which is expressed as
\begin{equation}
    \rm \mathbf{W}^c = Linear(cat(\mathbf{P}, \mathbf{F}))
\end{equation}
where $\rm \mathbf{W}$ is the learned weights by the point coordinates and point features for the aggregation of the following offset features. The weights give different sampled cylinder features different degrees of attention to achieve different degrees of feature aggregation. 
With the learned weights and sampled cylinder features, a feature aggregation approach is performed as
\begin{equation}
    \rm \mathbf{F}_{a} = Linear(\mathbf{F} + \sum_{k_c=1}^{K_c}\mathbf{W}^{c}_{k_c}\cdot \mathbf{F}^c_{o;k_c})
\end{equation}
where $\rm \mathbf{F}_{a}$ is the point feature after aggregating the cylinder features which represent the latitude and longitude properties. 

\textbf{Multi-Nearest Neighbor Feature Learning (MFL).}
From the perspective of the cylinder map, we have realized the modeling and learning of latitude and longitude features. For the modeling of geometric features inside point sets, we propose a Multi-nearest neighbor Feature Learning method called MFL. Specifically, we learn point cloud features by constructing neighboring points of different scales and then perform feature aggregation to realize geometric modeling of point cloud features of different scales, as shown in the top of Fig.~\ref{Fig3-CyDConv}.
\begin{figure*}[ht]
    \setlength{\abovecaptionskip}{1pt}
    \centering
    \includegraphics[width=1.0\linewidth]{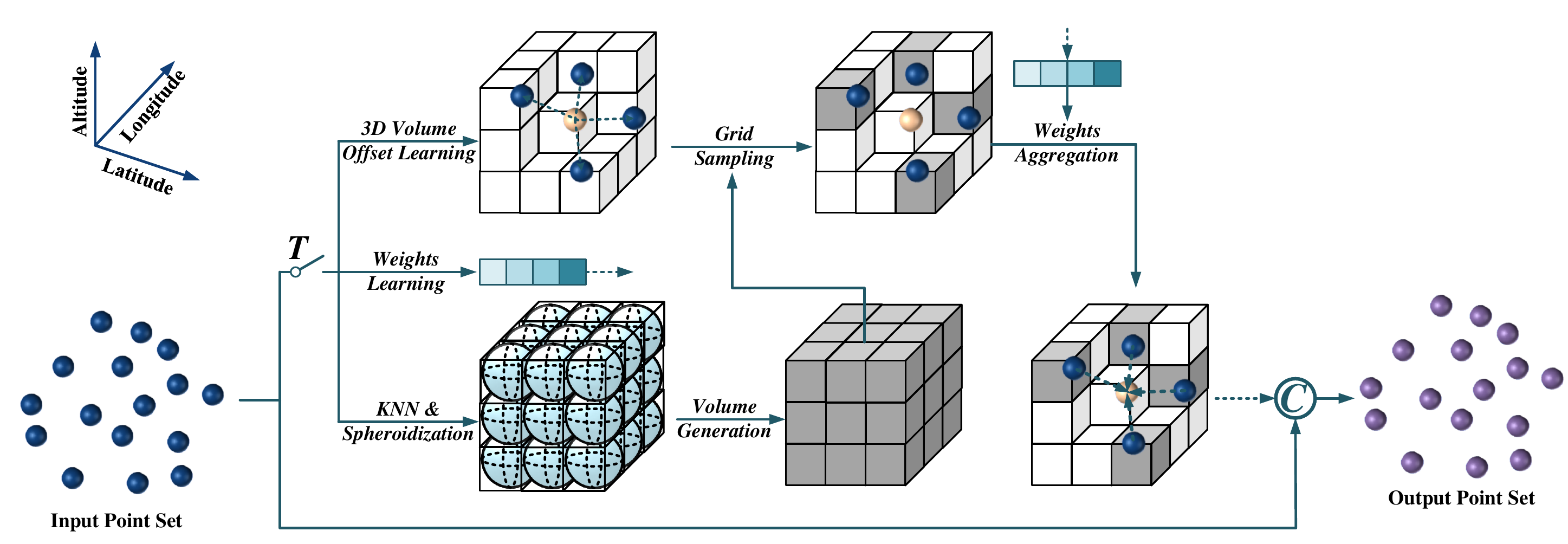}
    \caption{The architecture diagram of the proposed Sphere-wise Deformable point Convolution (SpDConv). `T' is a switch for whether to use SpDConv, which determines which connection path between the encoder and decoder uses SpDConv. Only the highest resolution connection path is selected to embed SpDConv in our network.}
    \label{Fig4-SpDConv}
\end{figure*}   

First, we construct neighbor search spaces of different scales which can be expressed as
\begin{equation}
    \rm\{\mathbf{T}_1, \mathbf{T}_2, \mathbf{T}_3 \}= \{KNN_1(\mathbf{P}), KNN_2(\mathbf{P}), KNN_3(\mathbf{P}) \}
\end{equation}
where $\rm \mathbf{T}_i$ is the neighbor search space of the point sets. $\rm KNN_i$ represents the k-nearest neighbor operation. 

\begin{equation}
    \rm \mathbf{F}^{N}_i = Linear(Grouping(\mathbf{T}_i, \mathbf{F}))
\end{equation}
where $\rm \mathbf{F}^{N}_i$ is the feature after the encoding and grouping based on the different neighbor search spaces. $\rm Grouping$ is the grouping layer proposed by PointNet++\cite{qi2017pointnet++}. 

Finally, we combine the obtained features $\rm \mathbf{F}^{N}_i$ based on different search spaces with the previously learned features $\rm \mathbf{F}_a$ representing latitude and longitude. 
\begin{equation}
    \rm \mathbf{F}^{CyDConv}_{out} = ReLU(BN(Linear(cat(\mathbf{F}^{N}_1, \mathbf{F}^{N}_2, \mathbf{F}^{N}_3, \mathbf{F}_a)))
\end{equation}
where $\rm ReLU$, $\rm BN$, $\rm Linear$, and $\rm cat$ represent the ReLU, Batch Normalization, linear layer, and concatenate operation, respectively. 
$\rm \mathbf{F}^{CyDConv}_{out}$ is the final output of point features in our CyDConv.

\subsection{Sphere-wise Deformable Point Convolution}
In addition to the latitude and longitude information, altitude information is a typical feature of ALS point clouds in geographic space arrangement. Unlike point clouds in visual scenes, altitude features are unique to ALS point clouds. To better model the characteristics of altitude, we propose Sphere-wise Deformable point Convolution (SpDConv) to adaptive learn altitude features. 

\textbf{Spheroidization and Volume Generation.} As shown in Fig.~\ref{Fig2-framework}, our proposed SpDConv is embedded in the feature propagation layer between the encoder and the decoder. Since the highest-resolution point set has the most comprehensive altitude features, our SpDConv only selects the highest-resolution feature aggregation layer for embedding.

From Fig.~\ref{Fig4-SpDConv}, given the highest resolution point set $\mathbf{P}\in \mathbb{R}^{\rm N\times 3}$ and point feature $\mathbf{F}\in \mathbb{R}^{\rm N\times M}$. We first choose whether to close switch `T'. When switch `T' is closed, SpDConv starts to perform operations. 
If the switch `T' is closed, unlike CyDConv, we first build a three-dimensional grid structure. The size of the grid structure is $\rm H_s$,$\rm W_s$, and $\rm Z_s$, which is expressed by the formula
\begin{equation}
    \rm \mathbf{S}_{cet} = GridArrange3D(H_s, W_s, Z_s)
\end{equation}
where $\rm \mathbf{S}_{cet}$ represents the center points of the three-dimensional grid constructed. 
$\rm GridArrange3D$ is the 3D Grid Arrangement operation. 
On this basis, we perform feature sampling, grouping, and aggregation operations on the original point cloud. Unlike CyDConv, SpDConv searches within each three-dimensional grid constructed to construct the feature aggregation range of the sphere. 
In each search sphere, we first determine the search radius, and then form a sub-point set as a neighbor point sphere set of the $\rm j$-th sphere center point by searching all points located in the sphere, which can be formulated as
\begin{equation}
    \rm \mathbf{\Psi }_{j} = \rm \bigcup \{\mathbf{F}_i\},\ \ s.t.\ ||\mathbf{P}_i-\mathbf{S}_{cet;j}|| < R_{s}
\end{equation}
where $\rm \mathbf{\Psi }_{j}$ demotes the set of points constructed with point $\rm \mathbf{S}_{cet;j}$ as the center of the sphere and falling within the radius $\rm R_{s}$.
Inside each sphere, we averaged the points of each set separately to obtain each element of the volume we constructed, which can be formulated as
\begin{equation}
    \rm \mathbf{V}_{j} = mean(\mathbf{\Psi }_{j})
\end{equation}
where $\rm \mathbf{V}_j$ denotes the volumn generated after spheroidization. The feature dimension of $\rm \mathbf{V}_j$ is $\rm H_s\times W_s\times Z_s$. The spheroidization operation performs feature learning on the three-dimensional grid arrangement of point clouds in three-dimensional space and finally outputs it in the form of the volume.

\textbf{Volume Offset Learning.} 
In order to allow the central point to better perceive the altitude features of neighboring points, it is necessary to perform different degrees of aggregation operations on the points in the surrounding area. 
Then, we generate the weight for the aggregation operation through the coordinates and features of the input point set, expressed as
\begin{equation}
    \rm \Delta^{s} = Linear(cat(\mathbf{P}, \mathbf{F}))
\end{equation}
where $\rm \Delta^{s}\in \mathbb{R}^{N_s\times K_s\times 3}$ denotes the offset learned by the input point set in 3D space. The offset generated provides the sampling position basis for subsequent feature sampling.

\textbf{Volume Feature Sampling and Aggregation.}
Given the sampling offset $\rm \Delta^{s}$ in 3D space, we first need to learn the aggregation weight of the sampling points in the 3D space based on the number of given offset points. 
The learning of the aggregation weight is also obtained by encoding the coordinates and features of the input point set through a linear layer, which can be expressed as
\begin{equation}
    \rm \mathbf{W}^s = Linear(cat(\mathbf{P}, \mathbf{F}))
\end{equation}
where $\rm \mathbf{W}^s$ denotes the aggregation weight, of which the dimension is $\rm N_s\times K_s$.

Given the sampling offset $\rm \Delta^{s}$ in 3D space, we adopt grid sampling in 3D space to sample the extracted volume features according to the position coordinates of the sampling points and aggregate them into the input point features, which can be formulated as
\begin{equation}
    \rm \mathbf{F}^s_{o} = GridSampling3D(\mathbf{V}, \mathbf{P}^s_{ref}+\Delta^{s})
\end{equation}
where $\rm \mathbf{F}^s_{o}$ denotes the output features after aggregation through grid sampling in 3D space. $\rm GridSampling3D$ represents the Grid Sampling operation in 3D space. $\rm \mathbf{P}^s_{ref}$ is the initial reference points of which the dimension is $\rm N_s\times 3$. 
    
With the learned aggregation weight $\rm \mathbf{W}^s$, we perform different degrees of aggregation operations on the sampled volume features to fully integrate the altitude information of surrounding point features into the central point feature, as
\begin{equation}
    \rm \mathbf{F}_s = \mathbf{F} + \sum_{{k_s}=1}^{{K_s}}\mathbf{W}^{s}_{k_s}\cdot \mathbf{F}^{s}_{o;k_s}
\end{equation}

At the same time, the high-resolution features $\rm \mathbf{F}_h$ from the encoder are fused with the obtained features after aggregation to obtain semantic features embedded with altitude information as
\begin{equation}
    \rm \mathbf{F}^{SpDConv}_{out} = ReLU(BN(Linear(Cat(\mathbf{F}_h, \mathbf{F}_s)))
\end{equation}
where $\rm \mathbf{F}^{SpDConv}_{out}$ is the features after the SpDConv operation. $\rm ReLU$, $\rm BN$, and $\rm Linear$ represent the ReLU function, Batch Normalization layer, and linear layer, respectively. $\rm Cat$ denotes the concatenate operation.

\subsection{Loss Function}
In the loss function design part of the network, we followed the loss function of RFFS-Net\cite{mao2022beyond} to drive our TDConvs for training. Specifically, we output the results of different resolutions respectively and then use the pre-generated point cloud labels of different resolutions as supervision information, which can be expressed as

\begin{equation}
\begin{aligned}
    \rm \mathcal{L} =&\rm \sum_{i=0}^{3}\lambda_i \mathcal{L}^{seg}_i\\
                =&\rm \sum_{i=0}^{3}\lambda_i \sum_{j=1}^{N_i}\sum_{c=1}^{C}[\mathcal{A}_i^{cj}log(S_{i}^{cj})+(1-\mathcal{A}_i^{cj})log(1-S_i^{cj})]
\end{aligned}    
\end{equation}
where $\rm\lambda_i$ represents the weights of our total loss function during training. In addition, $\rm\mathcal{A}_i$ is the supervised point set at different resolutions. $\rm S$ is the output results of different resolutions.

\section{Experiments}
\subsection{Experimental Setting}\label{sec:Experimental Setting}
\subsubsection{Datasets}\label{sec:Datasets} 
We selected the ISPRS Vaihingen 3D\cite{Cramer_2010,rottensteiner2012isprs,niemeyer2014contextual}, LASDU\cite{ye2020lasdu}, and DFC2019\cite{le20192019} datasets for experimental verification.

\textbf{ISPRS Vaihingen 3D:} 
The ISPRS Vaihingen 3D\cite{Cramer_2010,rottensteiner2012isprs,niemeyer2014contextual} dataset was collected with a Leica ALS50 system at an average altitude of 500m above Vaihingen, Germany. 
The point density of the dataset is approximately 6.7 points/m$^2$, and the feature of each point is composed of XYZ coordinates, reflectivity, return count information, and labels. 
ISPRS Vaihingen 3D is composed of nine categories: Powerline, Low Vegetation, Impervious Surfaces, Car, Fence/Hedge, Roof, Facade, Shrub, and Tree. 
Following the setting of the ISPRS Benchmark on 3D Semantic Labeling [13], the entire dataset is divided into two parts. 

\textbf{LASDU:}
LASDU\cite{ye2020lasdu} is a large-scale aerial LiDAR point cloud dataset acquired with an ALS system of type Leica ALS70 from an altitude of about 1200m over the valley along the Heihe River in the northwest of China. 
The dataset covers an urban area with approximately 1km$^2$ of highly dense residential and industrial buildings, including about 3.12 million points. 
The average point density is specified as approximately 3-4 pts/m$^2$. The semantic categories of LASDU are composed of Ground, Buildings, Trees, Low Vegetation, and Artifacts. 

\textbf{DFC2019:}
The DFC2019 dataset\cite{le20192019} is a widely used ALS large-scale point cloud data benchmark in the remote sensing field. The data benchmark was collected from two areas in Jacksonville and Omaha, USA, covering an area of about 100 square kilometers. Each point in the dataset contains five main features, namely xyz coordinates, intensity, and return number. In addition, as a semantic segmentation dataset, its categories consist of about 200 million points in 5 categories.

\subsubsection{Implementation Details}\label{sec:Implementation Details}
In the experiment, we adopted the PyTorch framework to complete our method. 
Specifically, for the data preprocessing process, we divide the point cloud of the datasets into patches according to a certain distance step (ISPRS Vaihingen 3D: $30m\times 30m$, LASDU: $50m\times 50m$, and DFC2019: $70m\times 70m$). 
For the training process, we randomly sample 4096 points from each patch as the input point sets of our TDConvs. 
In addition, the batch size and learning rate are set as 4 and 0.0002, respectively.
We trained for 500 epochs on a single A40 GPU. 
What's more, we adopt the Adam optimizer to minimize our target loss function. 
During testing, we take all points as input and evaluate the output results. The weights of our loss function are set to $\{1.0, 2.0, 2.0, 2.0, 2.0\}$ for the ISPRS Vaihingen 3D dataset, $\{1.0, 5.0, 5.0, 5.0, 5.0\}$ for LASDU dataset, and $\{1.0, 5.0, 5.0, 5.0, 5.0\}$ for DFC2019 dataset. 

In the input process of the network, the ISPRS Vaihingen 3D, LASDU, and DFC2019 datasets all adopt coordinate information and feature information as input features of the point set. The feature information of the ISPRS Vaihingen 3D dataset contains normalized XYZ coordinates and intensity information. The feature information of the LASDU dataset contains normalized XYZ coordinates, RGB normalization information, and intensity information. The feature information of the DFC2019 dataset contains normalized XYZ coordinates and intensity information.

\subsubsection{Evaluation Metrics}\label{sec:Evaluation Metrics}
In the experiments, we adopt OA and mF1 as the evaluation metrics of our method. Specifically, both OA and mF1 are based on Precision and Recall.
Precision refers to the ratio of the number of samples that are accurately classified as the positive category to the number of all samples that are classified as the positive category, which means how many samples in the samples that are predicted to be the positive category are really positive, which can be expressed as 
\begin{equation}
    \rm Precision=\frac{TP}{TP+FP}
\end{equation}
where TP and FP represent the cases of true positive and false positive, respectively. 

Recall is the ratio of the number of samples that are classified as the positive category to the actual number of samples in the test data set, which means how many of the samples that should be classified as the positive category are correctly classified, which is formulated as
\begin{equation}
    \rm Recall=\frac{TP}{TP+FN}
\end{equation}
where TP and FN represent the cases of true positive and false negative, respectively. 

After defining Precision and Recall, the evaluation indicators OA and F1 can be expressed as
\begin{equation}
\begin{aligned}
    \rm OA=\frac{TP+TN}{TP+TN+FP+FN}
\end{aligned}
\end{equation}
\begin{equation}
    \rm F1 = 2*\frac{Precision*Recall}{Precision+Recall}
\end{equation}
What's more, mF1 are the averages of F1 for all categories.

\subsection{Performance Analysis}\label{sec:Performance Analysis}
In this section, we analyze our method from both quantitative and qualitative perspectives, proving the superiority of our TDConvs.

\subsubsection{Quantitative Analysis}\label{sec:Quantitative Analysis} 
We present the segmentation performance of our TDConvs on three datasets, ISPRS Vaihingen 3D, LASDU, and DFC2019, respectively. 

\begin{table*}
\small
\caption{Performance of our TDConvs and other methods on the ISPRS Vaihingen 3D dataset. `vision' means vision-based methods. `Remote Sensing' represents remote sensing-based methods. The first 9 columns of the table represent the F1 score of each category. OA: overall accuracy. mF1: mean F1 score. \textbf{Bold} indicates the first performance, and \underline{underline} indicates the second performance.}\label{tab1-isprsperformance}
\centering
\begin{tabular}{c|l|ccccccccc|cc} 
\toprule[1.5pt]
 & Method &powerline&low\_veg & imp\_surf & car & fence & roof & facade &shrub & tree & OA & mF1\\
\midrule
\multirow{9}{*}{\rotatebox{90}{Vision}} & DGCNN\cite{wang2019dynamic}           &44.6 &71.2 &81.8 &42.0 &11.8 &93.8 &64.3 &46.4 &81.7 &78.3 &59.7 \\
~ & ConvPoint\cite{boulch2020convpoint}   &58.8 &80.9 &90.7 &65.9 &34.3 &90.3 &52.4 &39.1 &77.0 &81.5 &65.5 \\
~ & PointNet++\cite{qi2017pointnet++}     &57.9 &79.6 &90.6 &66.1 &31.5 &91.6 &54.3 &41.6 &77.0 &81.2 &65.6 \\ 
~ & PointConv\cite{wu2019pointconv}       &65.5 &79.9 &88.5 &72.1 &25.0 &90.5 &54.2 &45.6 &75.8 &79.6 &66.3 \\
~ & PointSIFT\cite{jiang2018pointsift}    &55.7 &80.7 &90.9 &77.8 &30.5 &92.5 &56.9 &44.4 &79.6 &82.2 &67.7 \\
~ & KPConv\cite{thomas2019kpconv}         &63.1 &82.3 &91.4 &72.5 &25.2 &94.4 &60.3 &44.9 &81.2 &83.7 &68.4 \\
~ & PointCNN\cite{li2018pointcnn}         &61.5 &\underline{82.7} &\underline{91.8} &75.8 &35.9 &92.7 &57.8 &\underline{49.1} &78.1 &83.3 &69.5 \\ 
~ & SCF-Net\cite{fan2021scf}              &64.2 &81.5 &90.8 &73.9 &35.2 &93.6 &61.5 &43.4 &82.6 &83.2 &69.8 \\
~ & Randla-net\cite{hu2020randla}         &68.8 &82.1 &91.3 &76.6 &43.8 &91.1 &61.9 &45.2 &77.4 &82.1 &70.9 \\
~ & GA-Net\cite{deng2021ga}               &65.6 &\bf{83.3} &90.6 &77.1 &41.6 &93.4 &61.1 &46.9 &80.3 &82.9 &71.1 \\
\midrule
\multirow{12}{*}{\rotatebox{90}{Remote Sensing}} & UM\cite{horvat2016context}        &46.1 &79.0 &89.1 &47.7 &5.2 &92.0 &52.7 &40.9 &77.9 &80.8 &59.0 \\ 
~ & BIJ\_W\cite{wang2018deep}         &13.8 &78.5 &90.5 &56.4 &36.3 &92.2 &53.2 &43.3 &78.4 &81.5 &60.3 \\
~ & WhuY3\cite{yang2017convolutional} &37.1 &81.4 &90.1 &63.4 &23.9 &93.4 &47.5 &39.9 &78.0 &82.3 &61.6 \\ 
~ & RIT\_1\cite{yousefhussien2018multi}&37.5 &77.9 &91.5 &73.4 &18.0 &94.0 &49.3 &45.9 &82.5 &81.6 &63.3 \\ 
~ & LUH\cite{niemeyer2016hierarchical}&59.6 &77.5 &91.1 &73.1 &34.0 &94.2 &56.3 &46.6 &\underline{83.1} &81.6 &68.4 \\ 
~ & D-FCN\cite{wen2020directionally}  &70.4 &80.2 &91.4 &78.1 &37.0 &93.0 &60.5 &46.0 &79.4 &82.2 &70.7 \\ 
~ & DANCE-NET\cite{li2020dance}       &68.4 &81.6 &\bf{92.8} &77.2 &38.6 &93.9 &60.2 &47.2 &81.4 &83.9 &71.2 \\
~ & GADH-Net\cite{li2020geometry1}     &\underline{75.4} &82.0 &91.6 &77.8 &44.2 &94.4 &61.5 &\bf{49.6} &82.6 &\bf{84.5} &\underline{73.2} \\
~ & VD-LAB\cite{li2022vd}    & 69.3 & 80.5 & 90.4 & 79.4 & 38.3 & 89.5 & 59.7 & 47.5 & 77.2 & 81.4 & 70.2 \\
~ & RFFS-Net\cite{mao2022beyond}      &\bf{75.5} &80.0 &90.5 &78.5 &\underline{45.5} &92.7 &57.9 &48.3 &75.7 &82.1 &71.6 \\
~ & IPCONV\cite{zhang2023ipconv} & 66.8 & 82.1 & 91.4 & 74.3 & 36.8 & \bf{94.8} & \bf{65.2} & 42.3 & 82.7 & \bf{84.5} & 70.7 \\
~ & MCFN\cite{zeng2023multi} & 74.5 & 82.3 & \underline{91.8} & \underline{79.0} & 37.5 & \underline{94.7} & 61.7 & 48.7 & \bf{83.3} & \underline{84.4} & 72.6 \\
\midrule
\rowcolor{gray!15}
 & \bf{TDConvs (ours)}            & 67.0 & 82.4 & 91.6 & \bf{84.7} & \bf{48.7} & 94.2 & \underline{63.3} & 46.9 & 81.7 & \bf{84.5} & \bf{73.4} \\
\bottomrule[1.5pt]
\end{tabular}
\end{table*}

\begin{table*}
\normalsize
\caption{
Performance comparison of our TDConvs with other methods on the LASDU dataset. \textbf{Bold} indicates the first performance and \underline{underline} indicates the second performance.}\label{tab2-lasduperformance}
\centering
\begin{tabular}{l|ccccc|cc}
\toprule[1.5pt]
Method& Ground & Buildings & Trees & Low Vegetation & Artifacts &OA &mF1 \\
\midrule
PointNet++\cite{qi2017pointnet++}         	&87.74 	&90.63 	&81.98 	&63.17 	&31.26 &82.84  &70.96 \\
PointCNN\cite{li2018pointcnn}  	       	&89.30 	&92.83 	&84.08 	&62.77 	&31.65 &85.04  &72.13 \\
DensePoint\cite{liu2019densepoint}	     	&89.78 	&94.77 	&85.20 	&65.45 	&34.17 &86.31  &73.87 \\
DGCNN\cite{wang2019dynamic}	        	&90.52 	&93.21 	&81.55 	&63.26 	&37.08 &85.51  &73.12 \\
KPConv\cite{thomas2019kpconv}	  	       	&89.12 	&93.43 	&83.22 	&59.70 	&31.85 &83.71  &71.47 \\
PosPool\cite{huang2020deep}  	         	&88.25 	&93.67 	&83.92 	&61.00 	&38.34 &83.52  &73.03 \\
HAD-PointNet++\cite{liu2020closer}      	&88.74 	&93.16 	&82.24 	&65.24 	&36.89 &84.37  &73.25 \\
PointConv\cite{wu2019pointconv}         	&89.57 	&94.31 	&84.59 	&67.51 	&36.41 &85.91  &74.48 \\
RFFS-Net\cite{mao2022beyond}               &\underline{90.92}  &95.35  &\bf{86.81}  &\bf{71.01}  &44.36 &87.12  &\underline{77.69} \\
IPCONV\cite{zhang2023ipconv} & 90.47 & \underline{96.26} & 85.75 & 59.58 & \underline{46.34} & 86.66 & 75.67 \\
MCFN\cite{zeng2023multi} & \bf{91.60} & \bf{96.70} & 85.90 & 67.10 & 43.80 & \bf{88.00} & 77.00 \\
\midrule
\rowcolor{gray!15}
\bf{TDConvs (ours)}   & 90.86  & 95.50 & \underline{86.66} & \underline{67.90} & \bf{48.33} & \underline{87.61} &  \bf{77.85} \\
\bottomrule[1.5pt]
\end{tabular}
\end{table*}

\textbf{ISPRS Vaihingen 3D:} 
As shown in Table~\ref{tab1-isprsperformance}, we present the performance comparison of our TDConvs with other point-based methods on the ISPRS Vaihingen 3D dataset. 
In general, our TDConvs achieve 84.5\% on OA and 73.4\% on mF1, respectively, surpassing all other remote sensing-based and vision-based methods, which strongly proves the superiority of our TDConvs. 
Specifically, our TDConvs outperforms the popular RFFS-Net\cite{mao2022beyond} by 2.4\% on OA and 1.8\% on mF1. 
OA and mF1 evaluate the overall accuracy and category accuracy of the model, respectively. 
Obviously, our TDConvs achieves the best performance for both OA and mF1, and significantly outperforms the SOTA method. 
For specific categories, our TDConvs achieves balanced performance on each category with small variance in performance, and there are no categories with extremely low performance.
This is because ALS point clouds are strictly arranged according to latitude, longitude, and altitude. 
Our TDConvs can achieve balanced segmentation performance. 
In particular, for categories of car, fence, and facade, our TDConvs achieves the best or second segmentation performance, which strongly demonstrates the superiority of our approach. 
\begin{table*}
\normalsize
\caption{
Performance comparison of our TDConvs with other methods on the DFC2019 dataset.
\textbf{Bold} indicates the first performance and \underline{underline} indicates the second performance.
}\label{tbl4dfc}
\centering
\begin{tabular}{l|ccccc|cc}
\toprule[1.5pt]
Method& Ground & Trees & Buildings & Water & Bridge &OA &mF1 \\
\midrule
PointNet++~\cite{qi2017pointnet++}      	&98.30 	&95.80 	&79.70 	&4.40 	&7.30   &92.70 	&57.10 \\
PointSIFT~\cite{jiang2018pointsift}       	&98.60 	&97.00 	&85.50 	&46.40 	&60.40  &94.00  &77.60 \\
PointCNN~\cite{li2018pointcnn}           	&98.70 	&97.20 	&84.90 	&44.10 	&65.30  &93.80  &78.00 \\
KPConv~\cite{thomas2019kpconv}              &98.40 	&94.20 	&87.40 	&43.00 	&77.50  &94.50  &80.10 \\
DGCNN~\cite{wang2019dynamic}                &97.88 	&93.22 	&90.37 	&88.23 	&54.39  &95.08  &84.82 \\
D-FCN~\cite{wen2020directionally}	        & 99.10   	&98.10 	&89.90 	&45.00 	&73.00  &95.60  &81.00 \\
DANCE-NET~\cite{li2020dance}	            & 99.10  	&93.90 	&87.00 	&58.30 	&83.90  &96.80  &84.40 \\
PointConv~\cite{wu2019pointconv}            &97.33 	&95.82 	&93.63  &74.50 	&69.24  &95.32  &86.10 \\
RFFS-Net~\cite{mao2022beyond}               &96.61  &96.10  &88.69  &77.84  &80.97  &94.31  &88.04  \\
LGGAA~\cite{jiang2022local}                 &98.90  &96.10  &90.20  &41.60  &83.70  &94.80  &81.40 \\
VD-LAB~\cite{li2022vd}                      &98.04  &95.53  &92.07  &85.51  &71.45  &95.93  &\underline{88.52} \\
IPCONV~\cite{zhang2023ipconv}               &98.80  &97.50  &92.90  &\underline{92.10}  &58.20  &97.10  &87.90 \\
LGENet~\cite{lin2021local}            &\textbf{99.30}  &\textbf{98.30}  &92.80  &47.40  &79.10  &\textbf{98.40}  &83.40 \\
DA-Net~\cite{zhang2022dual}           &\textbf{99.30}  &97.60  &92.70  &41.60  &\underline{85.10}  &\underline{98.30}  &83.30 \\
RRDAN~\cite{zeng2023recurrent}        & 99.10   &\underline{98.10}  &\underline{95.80}  &62.80  &82.30  &98.10  &87.60 \\
\midrule
\rowcolor{gray!15}
\bf{TDConvs (ours)}                         &\underline{99.17}  &97.64  &\textbf{96.84}  &\textbf{94.41}  &\textbf{86.80}  &98.24  &\textbf{94.97}   \\
\bottomrule[1.5pt]
\end{tabular}
\end{table*}

\begin{figure*}
	\setlength{\abovecaptionskip}{1pt}
	\centering
	\includegraphics[width=0.9\linewidth]{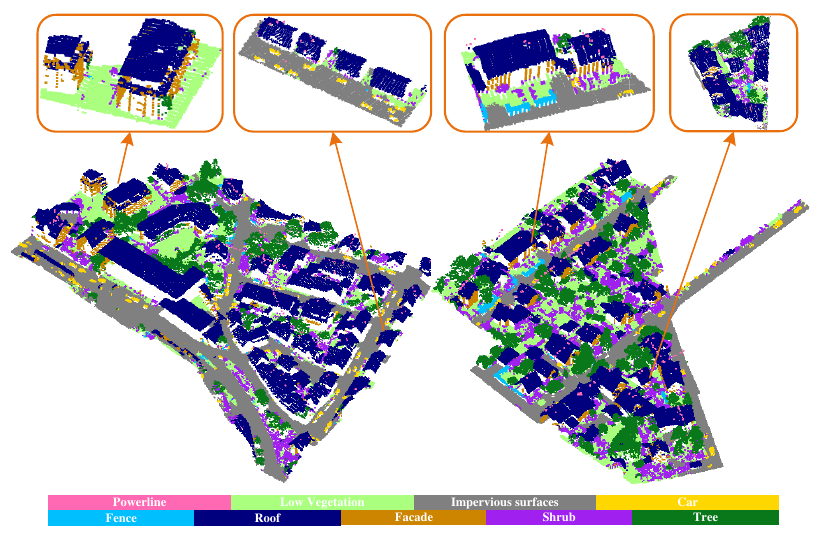}
	\caption{Segmentation results of the proposed TDConvs on the ISPRS Vaihingen 3D test set. The results in the box are local magnifications of the large area.}
	\label{Fig4-isprsvisual_all}
\end{figure*}
 
\textbf{LASDU:}
The comparison on the LASDU dataset of our TDConvs and other methods is given in Table~\ref{tab2-lasduperformance}. From the table, it can be concluded that our proposed TDConvs achieves the best segmentation performance, reaching an mF1 score of 77.85\%. In addition, on the OA indicator, our method is comparable to the optimal MCFN, but on the mF1 indicator, the TDConvs we proposed is 0.85\% higher than MCFN\cite{zeng2023multi}, which is a big improvement in ALS point cloud segmentation.
Compared with the popular RFFS-Net\cite{mao2022beyond}, our proposed TDConvs is 0.49\% and 0.16\% higher in OA and mF1 indicators respectively, which also reflects the superiority of the method in this paper.
Compared with OA, mF1 can better express the performance of point cloud segmentation because it takes into account the performance of each category, which is particularly obvious in datasets with imbalanced categories.
Therefore, in the field of segmentation, we focus more on the performance of mF1, and on the LASDU dataset, our TDConvs achieve the best mF1 performance, which strongly proves the superior performance of our TDConvs in the field of point cloud segmentation. 
As shown in Table~\ref{tab2-lasduperformance}, in specific categories, although our method does not achieve the best segmentation performance, the performance of each category is balanced. This has obvious advantages over IPCONV\cite{zhang2023ipconv}, which has large segmentation differences in categories. This is why our method can achieve the best mean F1 performance. 

\textbf{DFC2019:}
As shown in Table~\ref{tbl4dfc}, we give the performance comparisons of existing methods and our proposed TDConvs on the DFC2019 dataset. Obviously, our TDCOnvs achieves the best mF1 performance of 94.97\%, which is a high performance in the field of large-scale point cloud semantic segmentation. In addition, our method is 6.45\% higher than the existing SOTA method VD-LAB\cite{li2022vd}, which is a huge improvement. Through careful comparison, our TDConvs achieves the best performance in the Buildings, Water, and Bridge categories, the second best performance in the Ground category, and comparable performance to SOTA methods in the Trees category. This is because the TDConvs we proposed can explicitly model the latitude, longitude, and altitude information, which plays an important role in ensuring the balance of performance in each category. This is also the reason why our method can achieve superior performance in each category.

\subsubsection{Qualitative Analysis}\label{sec:Qualitative Analysis}
We also present the results of the qualitative analysis of TDConvs on the ISPRS Vaihingen 3D and LASDU datasets.

\textbf{ISPRS Vaihingen 3D:} 
From Fig.~\ref{Fig4-isprsvisual_all}, we provide qualitative visualization results of semantic segmentation of TDConvs on the ISPRS Vaihingen 3D test set. Through observation and analysis, our proposed TDConvs achieves superior results in overall segmentation results. Specifically, it can be found from the figure that the proposed TDConvs can perform better and accurate classification regardless of large-scale areas or small areas. In addition, in the locally enlarged area, especially small objects and objects with obvious height differences can be well classified. In addition, our TDConvs also has superior segmentation capabilities for sparse objects such as powerlines. 
\begin{figure*}
	\setlength{\abovecaptionskip}{1pt}
	\centering
	\includegraphics[width=1.0\linewidth]{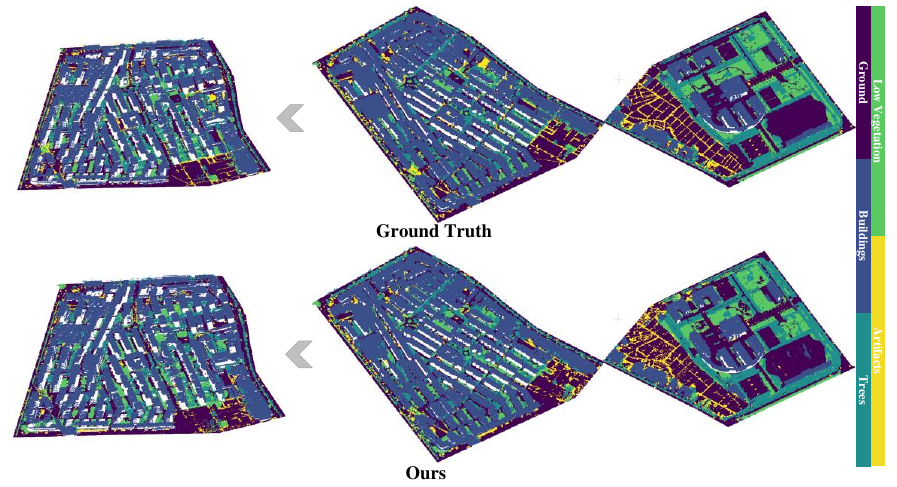}
	\caption{Segmentation results of the proposed TDConvs on the LASDU test set. The upper part is the ground truth of the dataset, and the lower part is the segmentation results of the proposed TDConvs.}
	\label{Fig5-visualLASDU_all}
\end{figure*}

\begin{figure}
	\setlength{\abovecaptionskip}{1pt}
	\centering
	\includegraphics[width=1.0\linewidth]{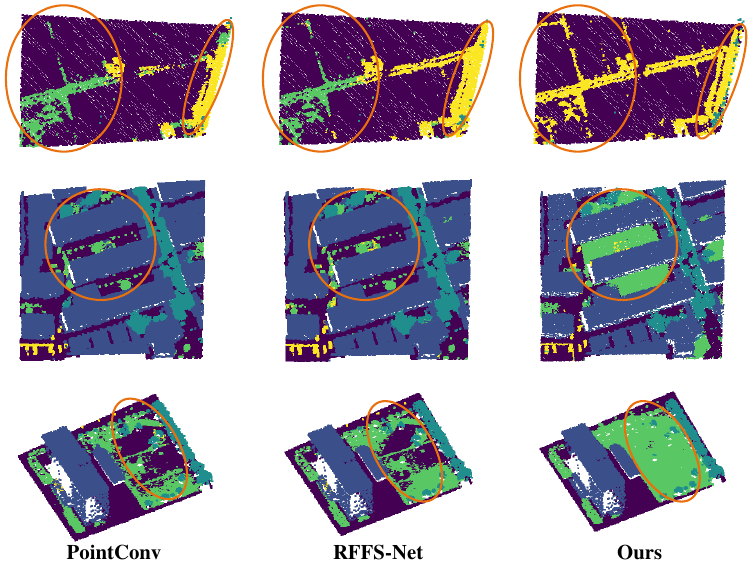}
	\caption{Comparison of the segmentation results of the proposed TDConvs with other methods on the LASDU test set. The first column is the segmentation results of PointConv\cite{wu2019pointconv}, the second column is the segmentation results of RFFS-Net\cite{mao2022beyond}, and the third column is the segmentation results of the proposed TDConvs.}
	\label{Fig6-visuallasdu_com}
\end{figure}
\begin{figure}
\setlength{\abovecaptionskip}{1pt}
\centering
\includegraphics[width=1.0\linewidth]{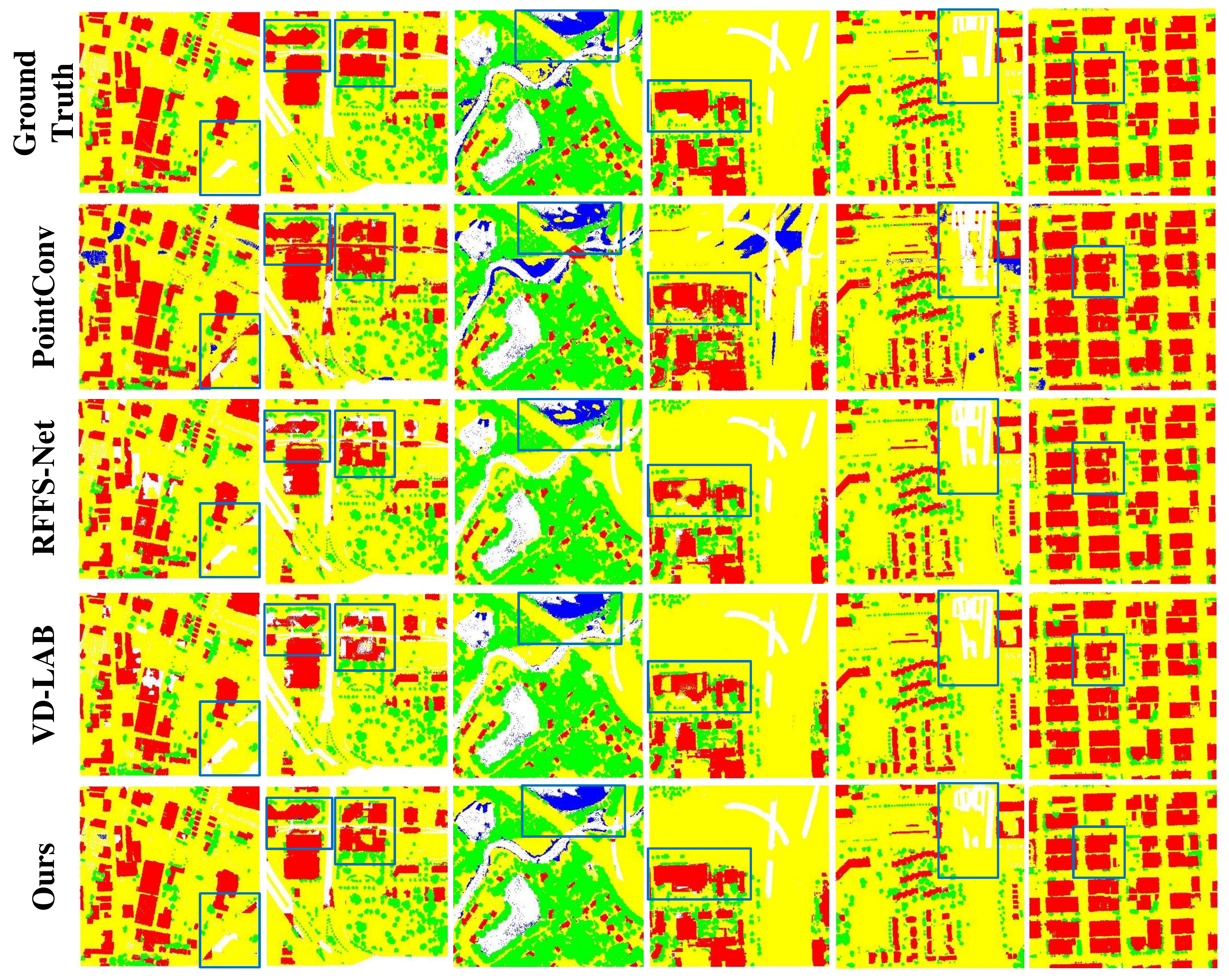}
\caption{
Comparison of the segmentation results of the proposed TDConvs with other methods on the DFC2019 test set. The first column is the Ground Truth, the second column is the segmentation results of PointConv\cite{wu2019pointconv}, the third column is the segmentation results of RFFS-Net\cite{mao2022beyond}, the fourth column is the segmentation results of VD-LAB\cite{li2022vd}, and the fifth column is the segmentation results of the proposed TDConvs.}
\label{Fig7-visualDFC2019}
\end{figure}

\textbf{LASDU:}
From Fig.~\ref{Fig5-visualLASDU_all}, we give the qualitative analysis results of semantic segmentation of TDConvs on the LASDU dataset. In comparison with Ground Truth (GT), the prediction results of TDConvs are basically consistent with GT and achieve superior segmentation results on objects such as roads and buildings. 
By changing the visual angle of part of the area, it can be found that buildings and trees with obvious height differences are well distinguished, which is challenging in the segmentation task of large-scale point cloud scenes.
\begin{table*}[ht]
\small
\setlength{\abovecaptionskip}{5pt}
\setlength{\belowcaptionskip}{10pt}
\caption{ 
Ablation experimental analysis of various components of the proposed TDConvs. 
(a) Performance of baseline method. 
(b) Performance of baseline which adds CyDConv without MFL. 
(c) Performance of baseline which adds CyDConv with MFL. 
(d) Performance of baseline with both CyDConv and SpDConv. 
`$\Delta$' represents the increment of performance or parameters.
} \label{tab3-ablationstudy}
\renewcommand\arraystretch{1.2}
\centering
\begin{tabular}{c|ccc|cccc|cc} 
\toprule[1.5pt]
 & CyDConv w/o MFL & CyDConv w/ MFL & SpDConv & OA & mF1 & $\Delta$(mF1) & $\sum\Delta$ & Params. & $\Delta$(Params.)  \\ 
\midrule
(a) &  \color{gray}\XSolidBrush   &  \color{gray}\XSolidBrush  & \color{gray}\XSolidBrush  & 82.2 & 68.6  & -    &  -   & 1.42M & - \\
(b) & \CheckmarkBold &  \color{gray}\XSolidBrush & \color{gray}\XSolidBrush & 83.6 & 70.8  & +2.2 &  2.2  & 2.24M & +0.82M \\
(c) & \CheckmarkBold & \CheckmarkBold & \color{gray}\XSolidBrush & 84.0 & 72.2  & +1.4 &  3.6  & 3.28M & +1.04M\\
(d) & \CheckmarkBold & \CheckmarkBold & \CheckmarkBold & \bf{84.5} & \bf{73.4} & +1.2 & \bf{4.8} &  3.31M & +0.03M \\
\bottomrule[1.5pt]
\end{tabular}
\end{table*}

In addition, we also compared the prediction results with other methods in Fig.~\ref{Fig6-visuallasdu_com}, including the popular PointConv\cite{wu2019pointconv} and RFFS-Net\cite{mao2022beyond}. Through comparison, it can be found that the TDConvs we proposed has achieved superior segmentation results in road segmentation, objects with large height differences such as buildings and trees, and highly similar objects such as the ground and low vegetation. However, PointConv\cite{wu2019pointconv} and RFFS-Net\cite{mao2022beyond} have poor performance in these two fields. In this case, there is no advantage for these two methods. The reason why our TDConvs achieve such superior results is that the proposed TDConvs can well model the latitude, longitude, and altitude characteristics of point clouds in large-scale scenes, which is greatly beneficial to the segmentation of different instance objects, especially for larger instances that are difficult to separate and with large height differences or high similarity.

\textbf{DFC2019:}
Fig.~\ref{Fig7-visualDFC2019} shows the visual comparisons of existing methods (including PointConv, RFFS-Net, and VD-LAB) and our method on the DFC2019 dataset. As can be seen from the figure, the method in this paper has superior segmentation performance compared to other methods, which is clearly shown in the blue boxes. In particular, in different areas, including building areas and natural feature areas, our TDConvs can achieve superior segmentation in details and is consistent with the Ground Truth. The reason is that the explicit modeling of longitude, latitude, and altitude in this paper can have good feature extraction capabilities for such large-scale point clouds to obtain the best segmentation effect.

\subsection{Ablation Studies}\label{sec:Ablation Studies}
In terms of quantitative analysis, we conduct ablation studies on our proposed components separately. Through data analysis, we prove the effectiveness of our TDConvs. Specifically, all our ablation experiments are performed on the popular ISPRS Vaihingen 3D dataset.

\subsubsection{The Effect of CyDConv}
CyDConv is proposed to explicitly model the arrangement characteristics of point clouds in longitude and latitude in large-scale scenes. 
As shown in Table~\ref{tab3-ablationstudy}, based on (a) baseline, we added CyDConv but did not introduce MFL operation, and the network performance improved from 82.2\% OA and 68.6\% mF1 to 83.6\% OA and 70.8\% mF1. mF1 increased by 2.2 percent over the baseline model, which is very advantageous in the semantic segmentation task of large-scale point clouds. This is because the CyDConv we proposed can provide explicit longitude and latitude modeling for point cloud data obtained from the aerial perspective, which helps to improve segmentation accuracy. 

\subsubsection{The Effect of MFL}
Multi-scale neighbor point feature extraction can simultaneously extract features of point sets in different neighbor areas of the center point, promoting the network's fine feature learning of objects of different scales. The introduction of MFL can realize this function well. As shown in Table~\ref{tab3-ablationstudy}, based on (b), we added MFL operation to CyDConv, which improved the network performance from the original 83.6\% OA and 70.8\% mF1 to the current 84.0\% OA and 72.2\% mF1, respectively. OA and mF1 increased by 0.4\% and 1.4\%. With MFL, the network can extract features from various instances at different scales, promoting the improvement of the overall network segmentation performance. 
The reason is that the selection of different neighboring areas allows the center point to have different receptive fields, which is helpful for feature learning of objects of multiple scales in large-scale scenes.

\subsubsection{The Effect of SpDConv}
For the large-scale point cloud in remote sensing, latitude, longitude, and altitude are typical geographical information features. 
CyDConv can implement explicit modeling of latitude and longitude information, while SpDConv is responsible for mathematical modeling of instance altitude. The introduction of SpDConv can effectively learn the distribution of instances in altitude and improve the fine segmentation of objects with large height differences. 
SyDConv is introduced on the basis of (c), and the network performance increased from 84.0\% OA and 72.2\% mF1 in (c) to the current 84.5\% OA and 73.4\% mF1, which increased by 0.5\% and 1.2\% respectively. This is a large margin in the point cloud semantic segmentation task. The reason is that explicit altitude modeling can promote the network to accurately learn the features of objects with large height differences or highly similarity, which helps improve the performance of the overall network.
\begin{table*}[ht]
\small
\setlength{\abovecaptionskip}{5pt}
\setlength{\belowcaptionskip}{10pt}
\caption{Verification experiments on the grid scale of CyDConv and SpDConv. 
The first four columns represent the resolution scales of maps at different levels, and the fifth column represents the resolution scale of the volume (remaining unchanged).} \label{tab4-map}
\renewcommand\arraystretch{1.2}
\centering
\begin{tabular}{cccc|c|cc} 
\toprule[1.5pt]
Map L1 & Map L2 & Map L3 & Map L4 & Volume & OA & mF1 \\ 
\midrule
20$\times $20 & 10$\times $10 & 5$\times $5 & 5$\times $5 & 40$\times $40$\times $5 & 81.7 & 71.0 \\
40$\times $40 & 20$\times $20 & 5$\times $5 & 5$\times $5 & 40$\times $40$\times $5 & 82.7 & 70.7  \\
40$\times $40 & 20$\times $20 & 10$\times $10 & 5$\times $5 & 40$\times $40$\times $5 & \bf{84.5} & \bf{73.4} \\
40$\times $40 & 20$\times $20 & 10$\times $10 & 10$\times $10 & 40$\times $40$\times $5 & 83.8 & 72.4  \\
80$\times $80 & 40$\times $40 & 10$\times $10 & 10$\times $10 & 40$\times $40$\times $5 & 81.1 & 71.4  \\
80$\times $80 & 40$\times $40 & 20$\times $20 & 10$\times $10 & 40$\times $40$\times $5 & 81.7 & 70.7 \\
80$\times $80 & 40$\times $40 & 20$\times $20 & 20$\times $20 & 40$\times $40$\times $5 & 81.6 & 70.8  \\
\bottomrule[1.5pt]
\end{tabular}
\end{table*} 

\subsection{Deep Analysis}
To better study and demonstrate our proposed TDConvs, we conduct in-depth experiments and analyze them from different perspectives.

\subsubsection{Grid Scale of CyDConv and SpDConv}
Since the proposed CyDConv and SpDConv are both based on map generation and volume generation in small resolution, the selection of the resolution is a significant process that determines network performance. 
From Table~\ref{tab4-map}, we explore the impact of CyDConv's different resolution maps on overall performance by changing the resolution scales of maps at different levels.
Under the premise that the volume resolution remains unchanged, different map resolutions have different effects on the segmentation performance of our TDConvs. In particular, as the resolution of each layer is continuously increased, the segmentation performance shows a trend of first increasing and then decreasing. 
Excessive resolution does not bring better performance, but increases the amount of calculation and parameters. 
Therefore, TDConvs finally chose $[40\times 40, 20\times 20, 10\times 10, 5\times 5]$.
In addition, from Table~\ref{tab5-volume}, while ensuring that the resolution of the map remains unchanged, changing the resolution of the volume will also have different effects on the segmentation performance of our TDConvs. 
The increase in volume resolution causes a downward trend in performance, so we finally choose $[40\times 40\times 5]$ as the volume resolution.

\subsubsection{Sampling Number of CyDConv and SpDConv}
\begin{table}[htb]
\caption{Analysis of the number of offset points in 2D map and 3D volume. `offsets in 2D map' and `offsets in 3D volume' represent the number of offsets in the 2D map space and the number of offsets in the 3D volume space, respectively.} \label{tab6-offsets}
\centering
\begin{tabular}{cc|cc} 
\toprule[1.5pt]
offsets in 2D Map & offsets in 3D Volume & OA & mF1 \\ 
\midrule
2 & 4 & 81.6 & 71.6 \\
4 & 8 & \bf{84.5} & \bf{73.4}   \\
8 & 16 & 84.3 & 72.2 \\
16 & 32 & 82.2  & 71.3 \\
\bottomrule[1.5pt]
\end{tabular}
\end{table}
Within map and volume, different numbers of offset points have a certain impact on network performance.
We set different numbers of sampling points of CyDConv and SpDConv in the experiments to determine the parameters. 
As shown in Table~\ref{tab6-offsets}, the network can achieve optimal segmentation performance by setting the number of offsets in 2D map and the number of offsets in 3D volume to 4 and 8, respectively. 
The combination of 8 and 16 also achieved good performance, but it brought more calculations and the performance was lower than the combination of 4 and 8. 
An appropriate number of sampling points can promote CyDConv and SpDConv to better learn the longitude, latitude, and altitude information of surrounding points, which will help improve the overall network segmentation performance. 

\begin{table*}[htb]
\setlength{\abovecaptionskip}{5pt}
\setlength{\belowcaptionskip}{10pt}
\caption{Verification experiments on the grid scale of CyDConv and SpDConv. 
The first four columns represent the resolution scales of maps at different levels (remaining unchanged), and the fifth column represents the resolution scale of the volume.} \label{tab5-volume}
\centering
\resizebox{0.55\linewidth}{!}{
\begin{tabular}{cccc|c|cc} 
\toprule[1.5pt]
Map L1 & Map L2 & Map L3 & Map L4 & Volume & OA & mF1 \\ 
\midrule
40$\times $40 & 20$\times $20 & 10$\times $10 & 5$\times $5 & 40$\times $40$\times $5 & \bf{84.5} & \bf{73.4} \\
40$\times $40 & 20$\times $20 & 10$\times $10 & 5$\times $5 & 40$\times $40$\times $10 & 82.3 & 72.6 \\
40$\times $40 & 20$\times $20 & 10$\times $10 & 5$\times $5 & 40$\times $40$\times $20 & 81.2 & 71.3  \\
40$\times $40 & 20$\times $20 & 10$\times $10 & 5$\times $5 & 80$\times $80$\times $10 & 82.3 & 71.4 \\
40$\times $40 & 20$\times $20 & 10$\times $10 & 5$\times $5 & 80$\times $80$\times $20 & 81.4 & 71.2 \\
\bottomrule[1.5pt]
\end{tabular}}
\end{table*}
\subsubsection{Number of Nearest Neighbor Sampling Points in MFL}
\begin{table}[htb]
\caption{Combination analysis of the number of nearest neighbor points in the MFL module. $\rm T_1$, $\rm T_2$, and $\rm T_3$ represent the number of points in the three neighboring areas, respectively.} \label{tab7-samplingT}
\centering
\begin{tabular}{ccc|cc} 
\toprule[1.5pt]
$\rm T_1$ & $\rm T_2$ & $\rm T_3$ & OA & mF1 \\ 
\midrule
8 & 16 & 32 & 81.4 & 70.1 \\
16 & 32 & 64 & \bf{84.5} & \bf{73.4} \\
32 & 64 & 128 & 83.2 & 73.0  \\
\bottomrule[1.5pt]
\end{tabular}
\end{table}
The construction of neighboring points at different levels can promote the improvement of the network's perception of different areas. 
The introduction of MFL enables the network to learn neighbor features of different radii within the point set during the feature extraction process. 
To better determine the number of aggregation points for different neighbor points, we conducted ablation experiments on the number of different aggregation points. 
As shown in Table~\ref{tab7-samplingT}, we determine the final MFL neighbor point combination by setting different $\rm T_1$, $\rm T_2$, and $\rm T_3$ values. It is obvious that in the combination $[16,32,64]$, the network can make full use of points in different sensing areas to perform feature aggregation at different levels. Although the network combined as $[32,64,128]$ also achieved good performance, more sampling points brought unnecessary calculations.

\subsubsection{Paramaters Analysis}
\begin{figure}
	\setlength{\abovecaptionskip}{1pt}
	\centering
	\includegraphics[width=1.0\linewidth]{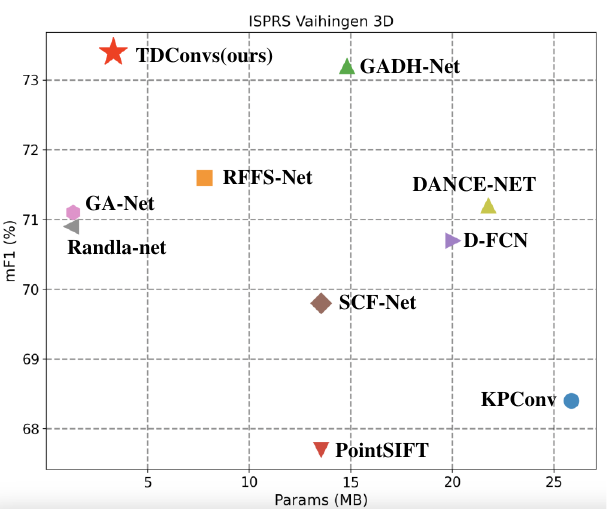}
	\caption{Parametric analysis of the proposed TDConvs. 
                The abscissa is the parameter amount, and the ordinate is the mF1 performance of segmentation. The closer the point is to the upper left corner, the more efficient the network is.}
	\label{Fig7-params}
\end{figure}

The parameter amount and performance of the network are important indicators to evaluate the efficiency of the model. 
In practical applications, low parameters and high segmentation performance are necessary. 
As shown in Fig.~\ref{Fig7-params}, we compared the number of parameters and segmentation performance of existing advanced methods and the proposed TDConvs and gave the relevant two-dimensional diagrams. 
The abscissa of the figure is the parameters of the models, and the ordinate is the mF1 indicator of segmentation. 
Obviously, the smaller the abscissa and the larger the ordinate represent the higher the efficiency of the model and the better the performance. 
It can be concluded from the figure that the TDConvs method we proposed is located in the upper left corner of the figure, which shows that we have achieved the best segmentation performance while ensuring the low number of parameters of our TDConvs, which effectively guarantees the practical application of lightweight.

As shown in Table~\ref{tab3-ablationstudy}, we give the calculation results of the number of parameters for (a) baseline, (b) baseline with CyDConv, (c) baseline with CyDConv and MFL, and (d) baseline with CyDConv, MFL, and SpDConv. From the given data, it can be seen that our TDConvs has a very large advantage in the overall number of parameters, and can achieve high segmentation performance while ensuring a low number of parameters. For each module, the increase in the number of parameters is within an acceptable range and is negligible. For CyDConv, method (b) increases the number of parameters by 0.82M compared to method (a). Furthermore, after adding MFL, the number of parameters increases from 2.24M in method (b) to 3.28M in method (c), an increase of 1.04M. Finally, SpDConv increases the number of parameters by 0.03M. These are tiny increases in the number of parameters and are within the acceptable range for actual network deployment. It can be seen that the method proposed in this paper ensures high segmentation performance while ensuring low parameter increase.

\subsubsection{Statistical Analysis}
\begin{figure}
	\setlength{\abovecaptionskip}{1pt}
	\centering
	\includegraphics[width=1.0\linewidth]{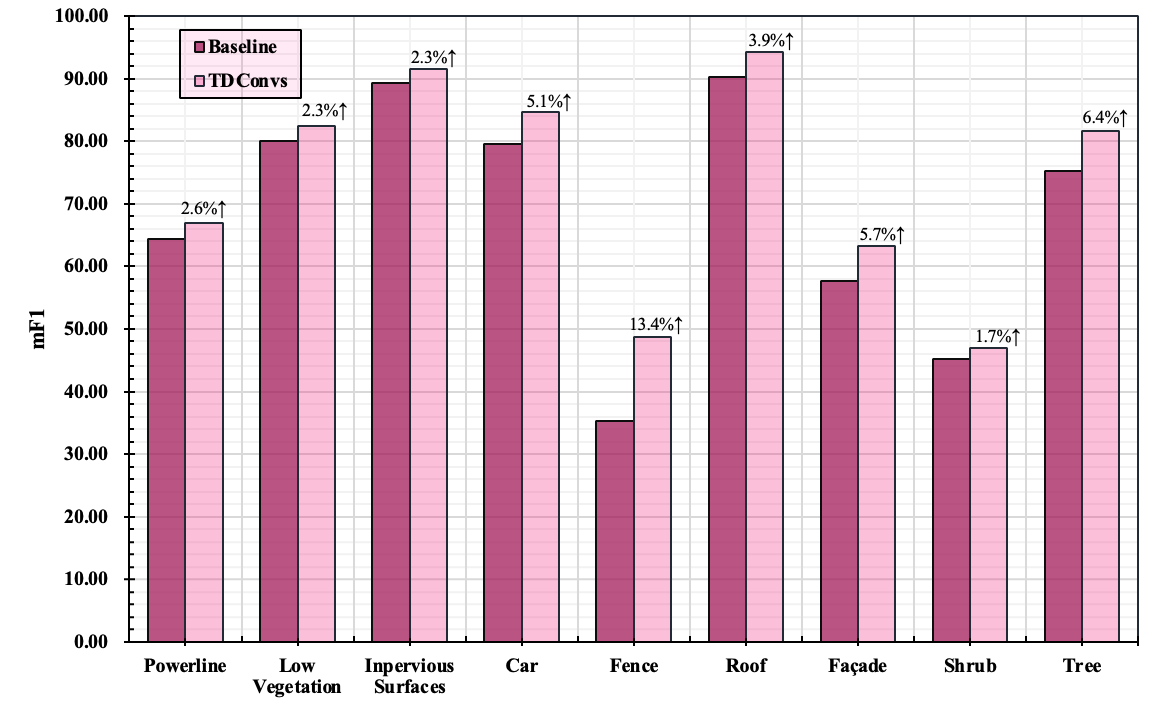}
	\caption{Category statistical analysis of the proposed TDConvs and baseline. 
                 The abscissa is each category, and the ordinate is the performance corresponding to each category.}
	\label{Fig8-Staanalysis}
\end{figure}

Analysis of statistical characteristics of network performance is necessary. 
As shown in Fig.~\ref{Fig8-Staanalysis}, we counted the category-by-category performance distribution of the baseline method and the TDConvs proposed in this paper. 
Through analysis, we found that the segmentation performance of our TDConvs is better than the baseline in all categories, and its performance is particularly outstanding in individual categories, such as the fence category and the tree category. 
From the statistical analysis, we can conclude that in point cloud segmentation of large-scale scenes, it is necessary to explicitly model its longitude, latitude, and altitude information, which can greatly promote the overall segmentation performance.

\section{Conclusion}\label{sec:Conclusion}
We proposed Twin Deformable point Convolutions (TDConvs) to explicitly model the geographic information characteristics of longitude, latitude, and altitude of large-scale aerial point clouds.
First, we proposed Cylinder-wise Deformable point Convolution (CyDConv), which performed adaptive aggregation of features by learning deformable sampling points of point clouds in a two-dimensional map space to improve the learning ability of longitude and latitude features. 
In addition, we introduced Sphere-wise Deformable point Convolution (SpDConv), which promoted the network for the fine-grained classification of highly different and highly similar instances through adaptively learning the altitude features of points in the three-dimensional volume space. 
By introducing TDConvs, the network efficiently learned geographical information features such as the latitude, longitude, and altitude information of point clouds, which helped in fine classification in actual large-scale scenarios and promoted implementation in practical applications. 
TDConvs provided a new perspective for the semantic segmentation task of large-scale point clouds in the remote sensing field.

\textbf{Future Work.} In future research, we will focus on two aspects: multi-source data fusion and foundation models of point cloud. First, the data of various existing sensors are developing rapidly. Single-modality data often has certain limitations and unreliable results for segmentation tasks, while the fusion of multi-source data can solve the unreliability of prediction results and improve the robustness of interpretation. Therefore, how to fuse point cloud data with other types of remote sensing data (such as remote sensing images) to improve the accuracy of data interpretation is the focus of our future research. In addition, the complexity and diversity of point cloud data require efficient and flexible processing methods. Traditional methods are usually effective for specific types of data or specific tasks, but difficult to extend to other scenarios. The large foundation model can learn general feature representations on large-scale and diverse data sets through pre-training, so it has a wider applicability. Therefore, how to build a point cloud foundation model under large-scale data is an urgent need for research in the field of remote sensing.

\section{Acknowledgments}

\bibliographystyle{cas-model2-names} 
\bibliography{egbib}  

%
%
%
\end{sloppypar}
\end{document}